\documentclass[letterpaper, 10 pt, conference]{ieeeconf}

\IEEEoverridecommandlockouts

\usepackage{graphicx}
\usepackage{subfig}
\usepackage{float}
\usepackage{amsmath}
\usepackage{amssymb}
\usepackage{color}
\usepackage{setspace}
\usepackage[hidelinks, colorlinks=true, linkcolor=black, urlcolor=blue, citecolor=black]{hyperref}
\usepackage[ruled,norelsize, linesnumbered]{algorithm2e}
\usepackage{tabularx, booktabs}
\usepackage{multirow}

\title{
    \LARGE \bf OpenAnnotate3D: Open-Vocabulary Auto-Labeling System for Multi-modal 3D Data
}

\author{Yijie Zhou$^1$, Likun Cai$^2$, Xianhui Cheng$^{1}$, Zhongxue Gan$^1$, Xiangyang Xue$^1$, and Wenchao Ding$^1$
\thanks{Corresponding author: Wenchao Ding and Xiangyang Xue. $^1$Fudan University, China. $^2$University of Toronto, Canada.\newline
{\tt\small Email: dingwenchao@fudan.edu.cn}
}
}

\begin{document}

\maketitle

\thispagestyle{empty}
\pagestyle{empty}

%%%%%%%%%% ABSTRACT

\begin{abstract}
In the era of big data and large models, automatic annotating functions for multi-modal data are of great significance for real-world AI-driven applications, such as autonomous driving and embodied AI.
Unlike traditional closed-set annotation, open-vocabulary annotation is essential to achieve human-level cognition capability. 
However, there are few open-vocabulary auto-labeling systems for multi-modal 3D data. In this paper, we introduce OpenAnnotate3D, an open-source open-vocabulary auto-labeling system that can automatically generate 2D masks, 3D masks, and 3D bounding box annotations for vision and point cloud data. Our system integrates the chain-of-thought capabilities of Large Language Models (LLMs) and the cross-modality capabilities of vision-language models (VLMs). To the best of our knowledge, OpenAnnotate3D is one of the pioneering works for open-vocabulary multi-modal 3D auto-labeling.
We conduct comprehensive evaluations on both public and in-house real-world datasets, which demonstrate that the system significantly improves annotation efficiency compared to manual annotation while providing accurate open-vocabulary auto-annotating results. 

\end{abstract}

\begin{source}
The source code will be released at \url{https://github.com/Fudan-ProjectTitan/OpenAnnotate3D}
\end{source}

% Leveraging the chain-of-thought capabilities of LLMs and the multi-modal representation capabilities of vision foundation models, our system can automatically generate open-vocabulary 2D and 3D segmentation annotations based on arbitrary high-level user-provided text descriptions, without requiring further training or fine-tuning. 
% Given any user-provided text description of semantic objects, we design a structured prompt and utilize LLMs to iteratively interpret the text to eliminate its prior knowledge until the subsequent vision module can comprehend the semantic information within it. 
% As a result, we can obtain 2D mask annotations for any object in the real world. 
% Additionally, after aligning the input 2D RGB and 3D point cloud data in the spatial dimension, we exhibit that generated 2D masks can be directly projected to the 3D point cloud space, serving as precise 3D object segmentation annotations. 
% To address multi-frame video data, we also employ a spatio temporal fusion and correction approach to refine single-frame annotations for higher precision. 

%%%%%%%%%% BODY TEXT

% fig1: 通过mask展示哪些类别是数据集内、开集的标签，其中凸显出大模型“知识”的特性和不局限于一个数据集的优势，caption里面尽量描述的清楚，把两者范式的关系写得像是一种发展的趋势和进化 

% fig2: 在流程图的基础上，把“知识库”和“只要标2D，3D自然就能对应上”的优势展示出来

% 其他figs，在需要表达某种文字不好轻松解释的概念里，把整体意思简图表达，配合文字说明和实际模型的参数来直观的表达&不失专业性，例如：画一个大脑，内有文字LLM，图中有概念示意图和模型实际输出参数的可视化

% 排版方面：图片的布局慢慢调整，概念要表达清楚，版面大小和图片的位置先大致定下来，文字尽量精简，赘述概念或者举例的部分通过图片和caption分摊空间

\section{INTRODUCTION}

The landscape of machine learning has been dominated by a paradigm where closed-set datasets are manually annotated for subsequent training and evaluation of learning models. 
A well-annotated benchmark can profoundly enhance the performance of corresponding tasks for both research and practical applications, exemplified by well-known datasets like ImageNet~\cite{deng2009imagenet}, COCO~\cite{lin2014microsoft}, KITTI~\cite{geiger2013vision}, and SemanticKITTI~\cite{behley2019semantickitti}. 

Data and annotations are undoubtedly the cornerstone of machine learning and deep learning tasks. 
Particularly, with the advent of Large Language Models (LLMs)~\cite{brown2020language, openai2023gpt4, chowdhery2022palm}, massive amounts of data have proven to lead to breakthrough improvements in model capabilities, as demonstrated by the emergence abilities of LLMs~\cite{wei2022emergent}. 
Compared with easily obtained textual corpora on the internet, which are used to train LLMs, acquiring well-annotated multi-modal (2D \& 3D) data is still a pending challenge.

\begin{figure}[t]
    \centering
    \includegraphics[width=\linewidth]{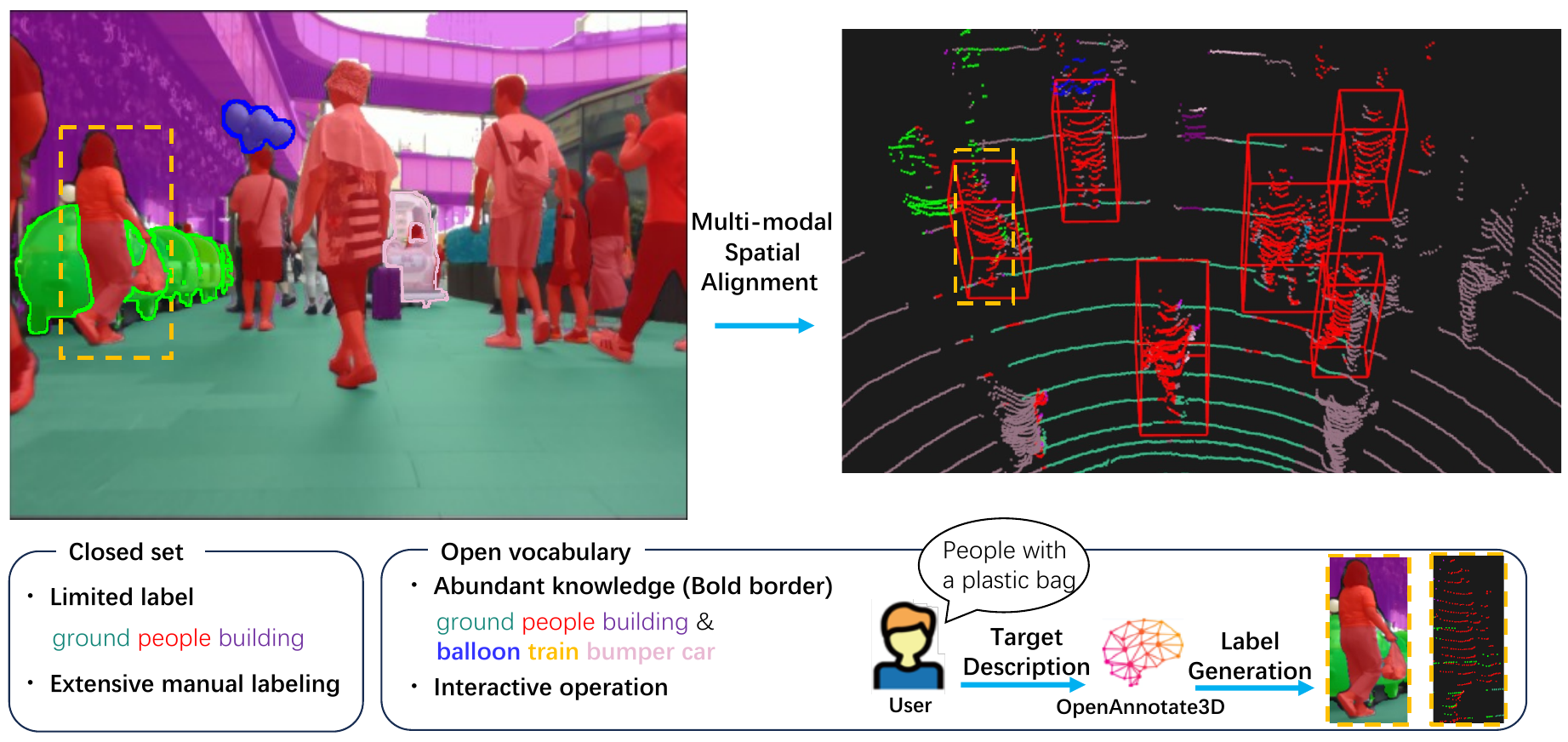}
    \caption{\small An illustration of open-vocabulary multi-modal 3D annotations. Compared to closed-set annotation systems which can provide labels for known categories such as ``\textit{ground}'', ``\textit{people}'' and ``\textit{building}'', OpenAnnotate3D can provide open-vocabulary 3D annotations for rare objects such as "\textit{balloon}" and "\textit{bumper car}". Moreover, OpenAnnotate3D can even understand high-level labeling commands such as "labeling the people with plastic bag".
    }
    \label{fig:ood_annotations}
    \vspace{-0.5cm}
\end{figure}

% The upper text box provides textual descriptions of various closed-set and open-set objects, accompanied by different colored blocks. 
% The lower images represent the results annotated by our OpenAnnotate3D, where different objects are segmented into various colors. 
% The bottom-left image displays the results of 2D segmentation, while the bottom-right image represents the effectiveness of 3D point cloud segmentation. 
% Our labeling system can easily handle various objects in both closed-set and open-set scenarios

Recently, the emergence of vision and language foundation models has underscored the urgency to develop an efficient annotation process for generating diverse and extensive multi-modal 3D datasets. Especially for applications like embodied AI and autonomous driving, huge amounts of annotations (2D \& 3D segmentation, 3D bounding boxes) are required. Moreover, unlike traditional closed-set data annotations, open-vocabulary scene understanding~\cite{peng2023openscene} is the common trend to enable human-level reasoning capability.
%Previous research~\cite{brown2020language, openai2023gpt4, kirillov2023segment, driess2023palm} has already revealed a direct relationship between data volume, model scale, and model capabilities. 
Manual annotation generation is significantly time-consuming and cannot satisfy the need for annotating open-vocabulary multi-modal 3D data. 
Consequently, there is a pressing need for \textit{an open-vocabulary auto-labeling tool that can automatically generate accurate 3D annotations for multi-modal data based on various user prompts}.

Regarding auto-labeling methods for multi-modal 3D data, there has not been an extensive amount of research in both academia and industry. 
Currently, one of the most advanced and effective approaches is the Auto-Labeling Machine showcased at Tesla's AI Day 2022~\cite{tesla_2022}, which is based on pre-trained models with closed-set taxonomy (predefined categories such as vehicles, pedestrians, lane topology, etc.). 
However, these pre-trained models struggle to effectively perform auto-labeling in open-vocabulary settings and fail to adapt to flexible labeling requirements.
%In an era of foundation models and embodied AI, there is still few open-vocabulary auto-labeling tool.

%For example, in practical application scenarios like autonomous driving, cameras and LiDAR sensors continuously capture multi-modal data in real time. 
% Moreover, this issue becomes even more pronounced when dealing with 3D scene data, especially 3D point clouds. 
% For most annotation tools supporting 3D data, users are required to perform a series of intricate operations within the 3D data space to complete the annotation process. 
% Additionally, due to the inherently complex and non-intuitive structure of 3D point clouds, annotations often need to be aligned with 2D RGB images together for quality evaluation.
% This demands annotators with a high degree of expertise to discern diverse objects across scenes and have rich annotation experience.
% This further intensifies the workload for annotators.

Recently, LLMs have demonstrated remarkable few-shot, zero-shot, and text reasoning capabilities across a range of natural language tasks, with the most notable application being ChatGPT~\cite{openai_chatgpt}. 
Taking inspiration from this, we propose a novel data annotation system called OpenAnnotate3D, which consists of an LLM-based interpreter module, a promptable vision module, and a spatio-temporal 3D auto-labeling process. 
Our annotation system, upon receiving multi-modal 3D data (vision and point clouds) and high-level labeling requests, such as ``\textit{labeling the balloon aside the road}'' and ``\textit{labeling the rightmost cyclist with a strange payload}''. The system explicitly reasons the request using the LLM-interpreter, automatically matches the textual information with specific objects in the semantic 3D world, and generates 2D mask, 3D mask, and 3D bounding box annotations as shown in Fig.~\ref{fig:ood_annotations}. There are two highlights for this system. First, the LLM-based interpreter module combines the LLM and promptable vision models (VLMs) in a closed-loop iterative manner, to interpret high-level user commands more precisely. Second, a spatio-temporal fusion and correction module is incorporated to overcome the imperfectness in single-frame results from VLMs.

Our contributions can be summarized as follows:
\begin{itemize}
\item A pioneering open-source open-vocabulary auto-labeling system for multi-modal 3D data.
\item An LLM-based interpreter that interacts with promptable vision modules in a closed-loop iterative manner enabling effective reasoning of high-level commands. 
\item A spatio-temporal fusion and correction method that overcomes imperfectness in single-frame auto-labeling.
\item Extensive experiments to validate the superior efficiency and open-vocabulary scene understanding capability of the proposed system.
\end{itemize}

\section{RELATED WORK}

\subsection{2D Annotation}

For annotating 2D RGB data, numerous tools have been developed like LabelMe~\cite{russell2008labelme}, Vatic~\cite{vondrick2013efficiently}, Label Studio~\cite{LabelStudio}, VIA~\cite{dutta2019vgg}, DEXTR~\cite{maninis2018deep}, PolygonRNN++~\cite{acuna2018efficient}, and CVAT~\cite{sekachevcomputer}.
These annotation tools cover most RGB-based vision tasks from basic image classification to video annotations. 
Since manual labeling is extremely time-consuming, most of these tools support model-assisted auto-labeling functions.
For example, DEXTR~\cite{maninis2018deep} and PolygonRNN++~\cite{acuna2018efficient} can be used to obtain precise dense annotations with manually provided coarse information, such as bounding boxes and extreme points.
Several open-sourced annotation tools like CVAT as well as commercial tools (Roboflow~\cite{Roboflow}, Labelbox~\cite{labelbox}) support SAM~\cite{kirillov2023segment} to boost the efficiency of annotating. However, most of these labeling tools remain in 2D and fail to handle multi-modal 3D data.

\begin{figure*}[t]
    \centering
	\includegraphics[width=0.85\linewidth]{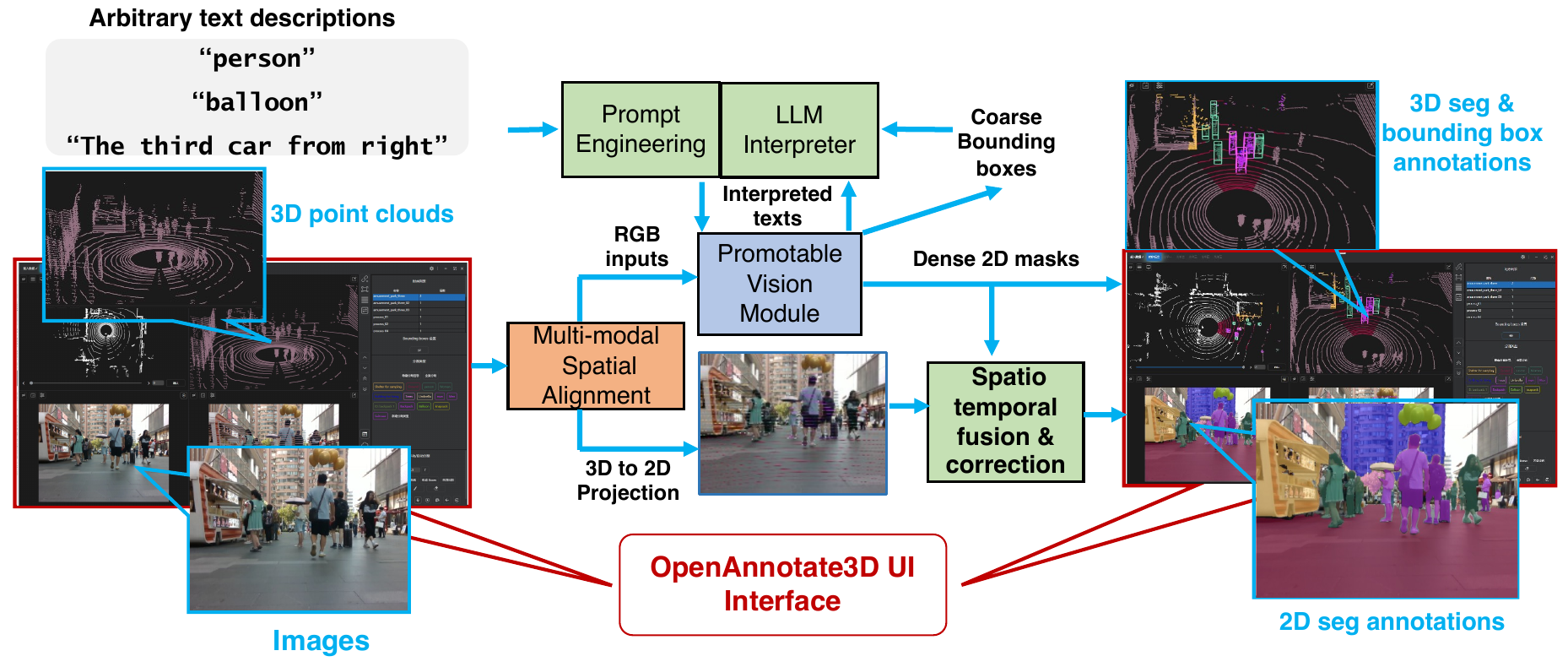}
    \caption{\small Workflow of OpenAnnotate3D. Upon receiving a user's labeling request, the system first reasons about the request through the LLM interpreter and proper prompt engineering. Note that the interpreter may interact with the promptable vision module for several rounds so that the interpreted texts fit the reasoning capability of the promptable vision module. Then dense 2D masks are produced and 3D masks are further calculated through multi-modal spatial alignment. To overcome the imperfectness in 2D masks, spatio-temporal fusion and correction are carried out to refine the 3D labels.
    }
    \label{fig:pipeline}
    \vspace{-0.3cm}
\end{figure*}

\subsection{3D Annotation}

Compared to annotating intuitive 2D RGB data, annotating 3D point clouds is inherently more complicated due to the sparsity and irregularity of 3D data. 
In open-sourced annotation tools mentioned above, only CVAT supports manual annotation of 3D bounding boxes on point clouds. 
In~\cite{golovinskiy2009min}, a min-cut base method was presented to segment a single object from the background in 3D point clouds. 
To expand to multi-object segmentation, \cite{monica2017multi} developed an interactive method based on the shortest path tree, requiring the user to select sparse control points in a 3D scene.
In~\cite{kontogianni2023interactive}, a deep network is introduced for 3D instance segmentation, which generalizes well to previously unknown objects with little manual annotation effort.
%3D BAT~\cite{zimmer20193d} supports semi-automatic 3D object detection annotation in multi-frame videos, based on manually provided annotations for the starting and ending frames.
If the input data includes both 2D RGB and 3D point clouds, LATTE~\cite{wang2019latte} and LiLaNet~\cite{piewak2018boosting} support 3D point cloud segmentation guided by 2D masks. PALF~\cite{zhang2023palf} uses a pre-trained 3D object detection model to generate 3D bounding boxes and calibrate them using 2D bounding boxes.

These annotation tools for 3D point clouds mentioned above generally either require users to annotate within the point cloud data space directly or have complex and intricate operational logic. 
All of these conditions significantly raise the threshold and workload for annotating 3D point clouds. Moreover, few of these annotation tools support open-vocabulary annotations. In contrast, OpenAnnotate3D provides a systematic solution for open-vocabulary auto-labeling for multi-modal 3D data.

\section{System Architecture}
\label{sec:system_arch}

In this section, we introduce the workflow of OpenAnnotate3D, as well as its implemented components in detail. 
Fig \ref{fig:pipeline} illustrates the whole auto-labeling process of our system, which takes a text description $T\in\mathbb{R}^{N}$, RGB image $I\in\mathbb{R}^{M\times N\times3}$, and 3D point clouds $P\in\mathbb{R}^{N\times 3}$ as input.
To further reduce the frequency of physical interaction for users, our system also supports voice input. 
These voice signals are automatically transcribed to text using a speech recognition model, Whisper~\cite{radford2023robust}.
Our system accomplishes the generation of precise 2D mask, 3D mask, and 3D bounding box annotations based on any user-provided descriptive text.
%In the following sections, we will provide a detailed module overview of our annotation system.

\begin{figure}[t]
    \centering
    \includegraphics[width=\linewidth]{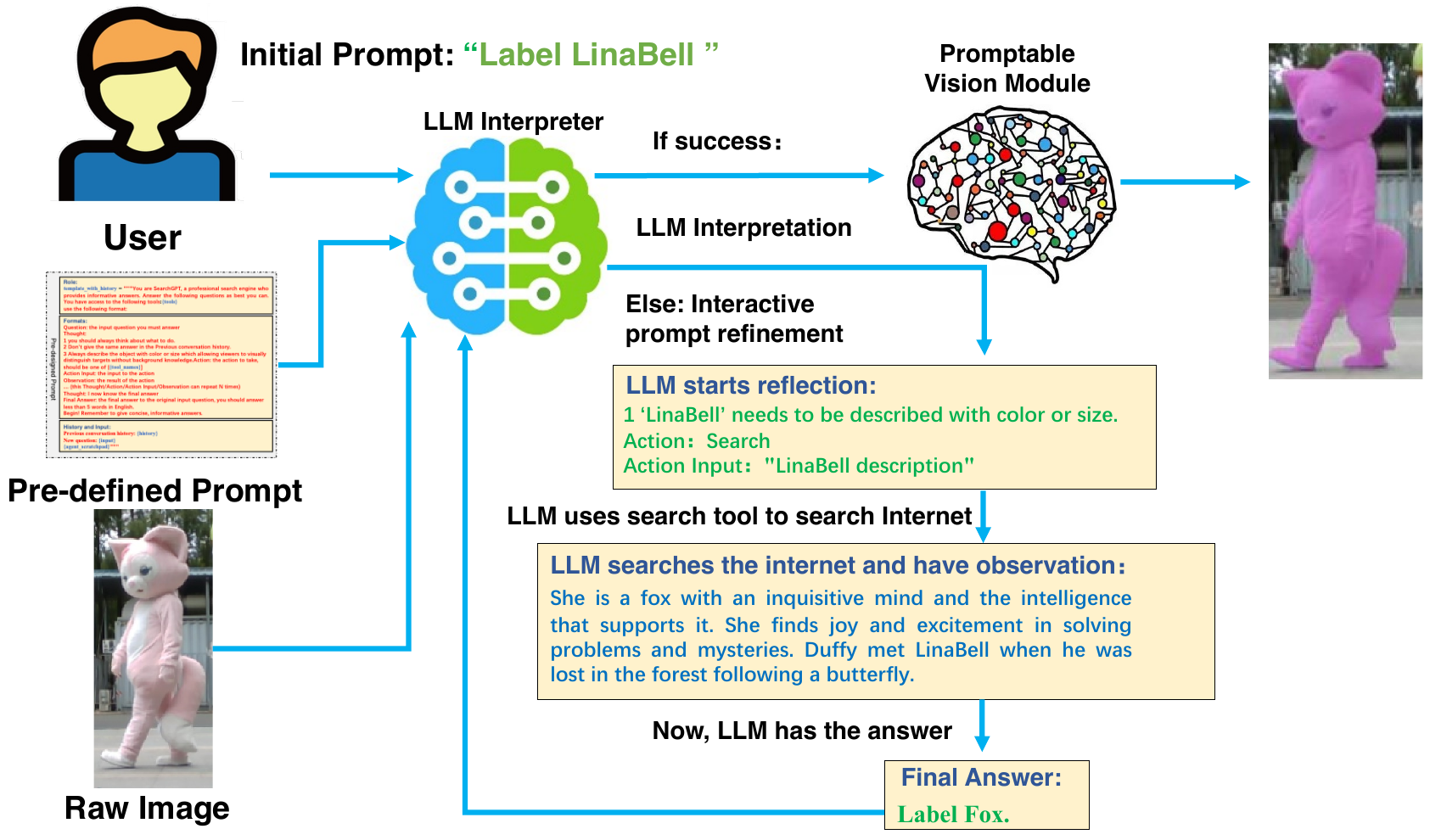}
    \caption{\small Illustration of the process of interpretation based on a pre-defined prompt. Using the pre-defined prompt template, a role can be assigned to the LLM, specifying available tools. Furthermore, interaction history with the promptable vision module is memorized and incorporated. }
    \label{fig:prompt}
    \vspace{-0.3cm}
\end{figure}

\subsection{LLM-based Interpreter Module} \label{subsec:llm}

%A pre-defined segmentation model is not suitable for open-vocabulary annotation, since it only understands a few keywords for a closed set of categories, which may fail to reason about diverse use inputs. 
Our system is designed to annotate one or multiple open-vocabulary instances based on flexible user-provided text descriptions. The labeling request can be high-level and abstract, such as ``labeling the balloon on the road". To this end, an LLM is employed as a semantic interpreter to transform the user-provided prompt into plain text outputs that can be understood by VLMs. 
The reason is that even the recent state-of-the-art promptable vision modules (VLMs) suffer from limited textual reasoning capabilities compared to LLMs, which may result in poor visual recognition results if we directly feed raw user text commands to the promptable vision module.
Given the LLM-based interpreter, users only need to provide a high-level text phrase for labeling commands rather than elaborately design a segmentation algorithm before the annotation process.

%In this section, we introduce how our annotation system processes user-provided text inputs using an LLM-based Interpreter module. For users of our system, we do not impose explicit formatting requirements on the prompt.

\begin{figure*}[t]
    \centering
    \includegraphics[width=0.8\linewidth]{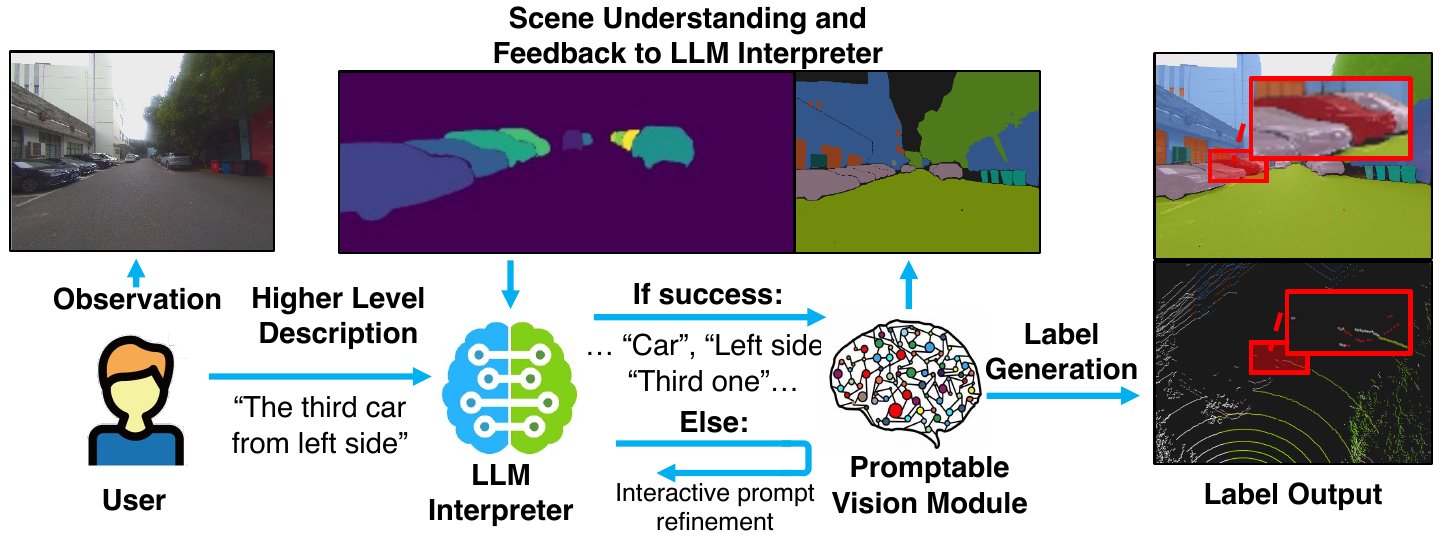}
    \caption{\small Pipeline for iterative text interpretation. LLM first interprets the user's goal prompt, extracts the core content, and then conducts an initial query to the promptable vision module to obtain scene features. The prompt for the vision module is then constantly refined by the LLM interpreter based on the scene understanding results from the promptable vision module. This enhances the reasoning capability and segmentation accuracy significantly.}
    \label{fig:long_prompt}
    \vspace{-0.5cm}
\end{figure*}

\subsubsection{\textbf{Prompt Engineering}}

For direct prompts such as ``\textit{garbage bin with trash on it}'', we augment the text with a pre-defined prompt, which is shown in Fig. \ref{fig:prompt}.  The prompt template primarily includes three components: 1) the fundamental role of the LLM interpreter and its basic task description; 2) several important rules for the interpreter regarding the format of outputs; 3) the conversation history of the last 5 text inputs from users. 
This prompt engineering enables LLMs to better interpret user-provided text, minimizing any prior knowledge, and thus allowing the subsequent vision module to achieve a higher hit rate.

For high-level prompts such as ``\textit{generate 3D bounding box for the third car from the left}'', we first parse user input using the LLM, extracting all relevant information (even from internet). Subsequently, we conduct a coarse query using the promptable vision module, deriving 2D segmentation data and feeding the segmentation quality back to the LLM. Drawing on its scene comprehension, the LLM can then accurately discern the user's intention and target, slightly adjust the output interpreted texts, and prompt the vision module again. A toy example is depicted in Fig.~\ref{fig:long_prompt}. This process can be iterative as elaborated as follows. 

\subsubsection{\textbf{Iterative Text Interpretation}}
We devise an iterative text interpretation strategy as outlined in Algo.~\ref{algo:iter_text_interpret}, which is designed to better connect open-vocabulary user prompts to a downstream promptable vision module. Initially, we feed the original user-provided text to the promptable vision module. If the vision module cannot establish a match between the text description and image, it provides feedback to the LLM interpreter. The prompt history is memorized and further incorporated into the next prompt. Then the LLM interpreter adjusts its outputs leveraging language understanding and reasoning abilities embedded in LLMs until the promptable vision module can understand its instructions well. 

Suppose the visual module still cannot generate a valid output after $L$ iterations, our annotation system interrupts and provides feedback to the user, requesting them to refine their text input to describe desired objects. Additionally, when mask annotations are generated, our system also allows users to assess these annotations. If they are dissatisfied with the results, this feedback is also conveyed to the interpreter, enabling the system to continue iterating for better annotations.

\setlength{\textfloatsep}{6pt}
\begin{algorithm}[t]
    \caption{Iterative Text Interpretation}
    \textbf{Inputs:} RGB image: $I$, user-provided text: $T_0$\;
    \textbf{Outputs:} A set of bounding boxes: $\mathcal{B}$\;
    $\mathcal{B} \leftarrow \varnothing$\;
    \For{$i \leftarrow 0$ to $L$}{
        $\mathcal{B} \leftarrow \mathtt{VisionModule}(T_i, I)$\;
        \eIf{$\mathcal{B} == \varnothing$}{
            $T_{i+1} \leftarrow \mathtt{LLMInterpreter}(\mathtt{PromptEng}(T_i))$\;
        }{
            End this for loop;
        }
    }
    \Return $\mathcal{B}$
    \label{algo:iter_text_interpret}
\end{algorithm}

\subsection{Promptable Vision Module and 3D Auto Labeling} \label{subsec:vision_module}

Following the LLM-based interpreter, we build a labeling process that can automatically annotate 3D multi-modal data. Current off-the-shelf cross-modality vision-language models are based on 2D images, such as CLIP~\cite{radford2021learning} and SAM~\cite{kirillov2023segment}. In this section, we will elaborate on how to annotate 3D multi-modal data based on off-the-shelf VLMs.

\subsubsection{\textbf{Multi-Modal Spatial Alignment}}
As aforementioned, our OpenAnnotate3D is designed to perform object-level labeling on RGB and 3D point cloud data. There are few open-vocabulary models directly operating on multi-modal 3D data. To this end, we conduct multi-modal spatial alignment so that the reasoning capability of 2D VLMs can be better utilized.

When RGB and 3D point clouds are spatially aligned, precise 2D masks can be directly projected to the 3D space to serve as 3D segmentation annotations. Now that 2D mask annotations are produced from the vision module, to obtain 3D annotations, we need to spatially align RGB camera images with 3D LiDAR point clouds.

We directly transform 3D point clouds in world coordinates into 2D image coordinates using the extrinsic and intrinsic parameters as follows:
\begin{equation}
    \mathbf{s}[u_i, v_i, 1]^\top = \overrightarrow{P}[x_i, y_i, z_i, 1]^\top
\end{equation}
where $[u_i, v_i, 1]^\top$ and $[x_i, y_i, z_i, 1]^\top$ are 2D homogeneous image coordinates and 3D homogeneous world coordinates, respectively. $\overrightarrow{P} = K\cdot[R~|~\mathbf{t}]$ represents the projection matrix, where $K\in\mathbb{R}^{3\times3}$ is the intrinsic matrix, and $[R~|~\mathbf{t}]\in\mathbb{R}^{3\times4}$ is the extrinsic matrix, consisting of the rotation matrix $\mathbb{R}$ and translation vector $\mathbf{t}$. $\mathbf{s}$ is a scaling factor.

\begin{figure}[tbp!]
    \centering
    \includegraphics[width=.9\linewidth]{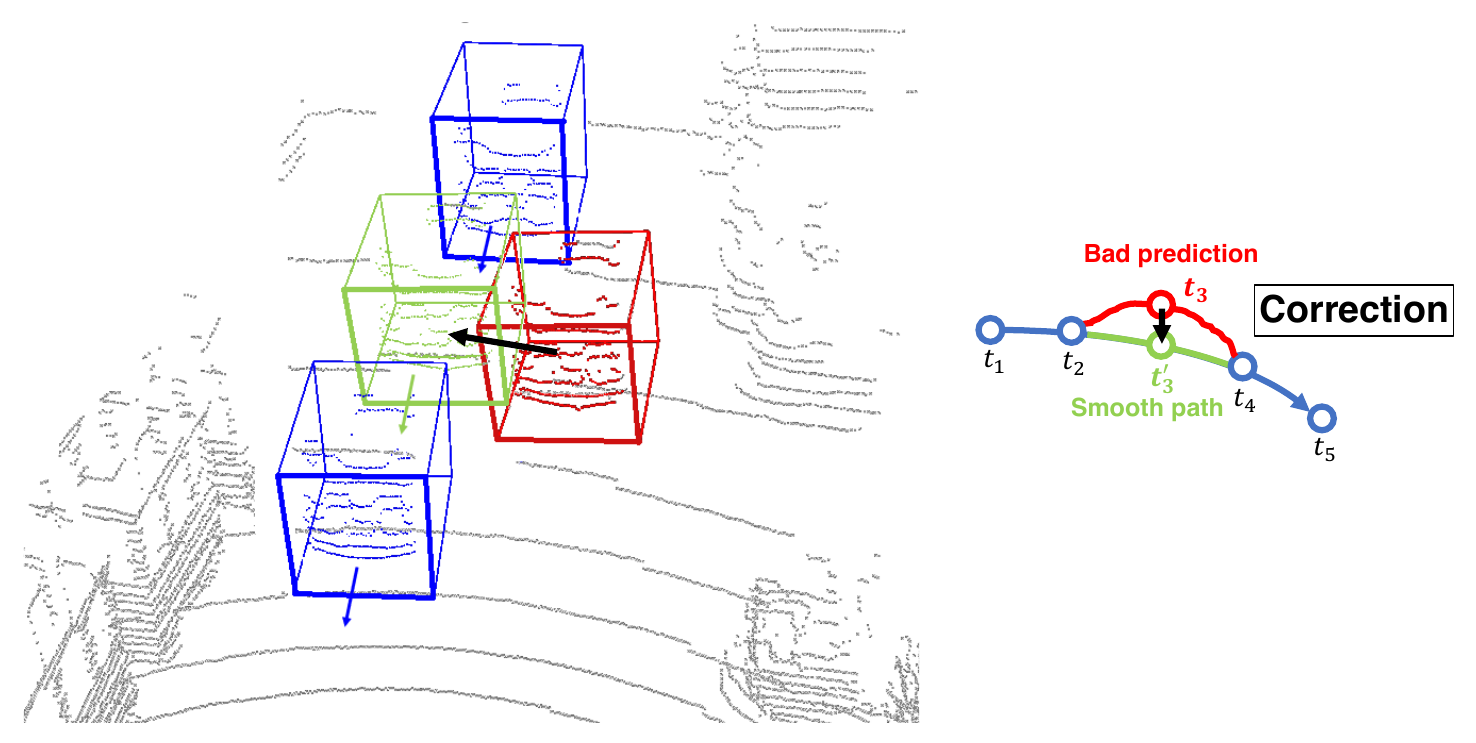}
    \caption{Illustration of spatio-temporal fusion and correction.}
    \label{fig:spatio_temporal_corridor}
\end{figure}

Given well-aligned RGB images and 3D point clouds, we can establish an accurate point-to-pixel correspondence. 
In Sec.~\ref{subsec:vision_module}, we obtain 2D masks through the promptable visual module, which is implemented using VLMs such as SAM. Interested readers may refer to Sec.~\ref{sec:exp} for implementation details. Based on the semantic object annotated in 2D image coordinates, we can label corresponding points within the same area as the same semantic object. When these point clouds are projected back to 3D world coordinates, we can directly obtain 3D mask annotations for different objects. Additionally, our system also supports labeling 3D bounding boxes by fitting 3D bounding boxes to segmented and clustered 3D point clouds.

\begin{figure*}[t]
    \centering
    \includegraphics[width=\linewidth]{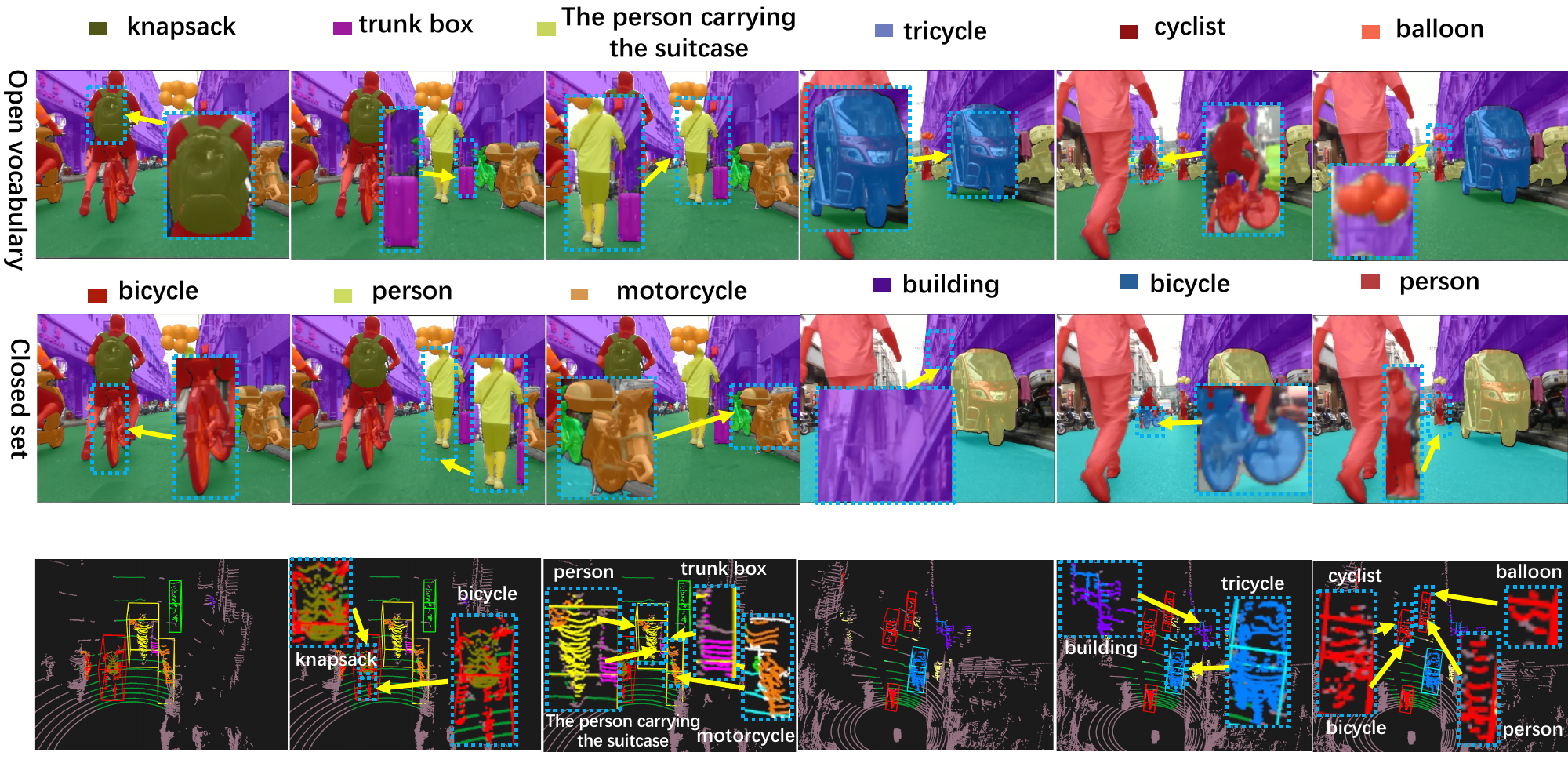}
    \caption{Visualization of open-vocabulary annotations generated by our OpenAnnotate3D in in-house datasets.}
    \label{fig:visualization}
    \vspace{-0.5cm}
\end{figure*}

\subsubsection{\textbf{Spatio-temporal Fusion and Correction}}

When dealing with multi-frame video data, we offer two optional solutions enabling continuous frame annotation. In the first approach, users can explicitly specify the starting and ending frames within a video segment. Once the system automatically labels the two frames, an interpolation algorithm is employed to annotate the remaining frames in this video. This approach is highly efficient but may not guarantee the accuracy of annotations for intermediate frames. 

Therefore, our system also supports frame-by-frame auto-labeling for videos. However, the issue is that the VLMs may occasionally mislabel or miss certain objects for particular frames, which may result in poor 3D annotation quality, especially for difficult cases such as occlusion. 

To this end, we propose a fusion and correction method based on the observation that it is essential to utilize spatial and temporal information across frames. If we consider time as an additional axis, a moving object will generate a 3-dimensional volume over time. A cross-section of this volume represents the object's instantaneous pose in time. Given that the majority of objects in the physical world adhere to kinematic laws, maintaining geometric and spatial consistency, we can evaluate and correct the trajectory of an object. Fig~\ref{fig:spatio_temporal_corridor} demonstrates how the spatio-temporal fusion and correction fix the result of an incorrect annotation.

\begin{figure}[tbp!]
    \centering
    \includegraphics[width=.9\linewidth]{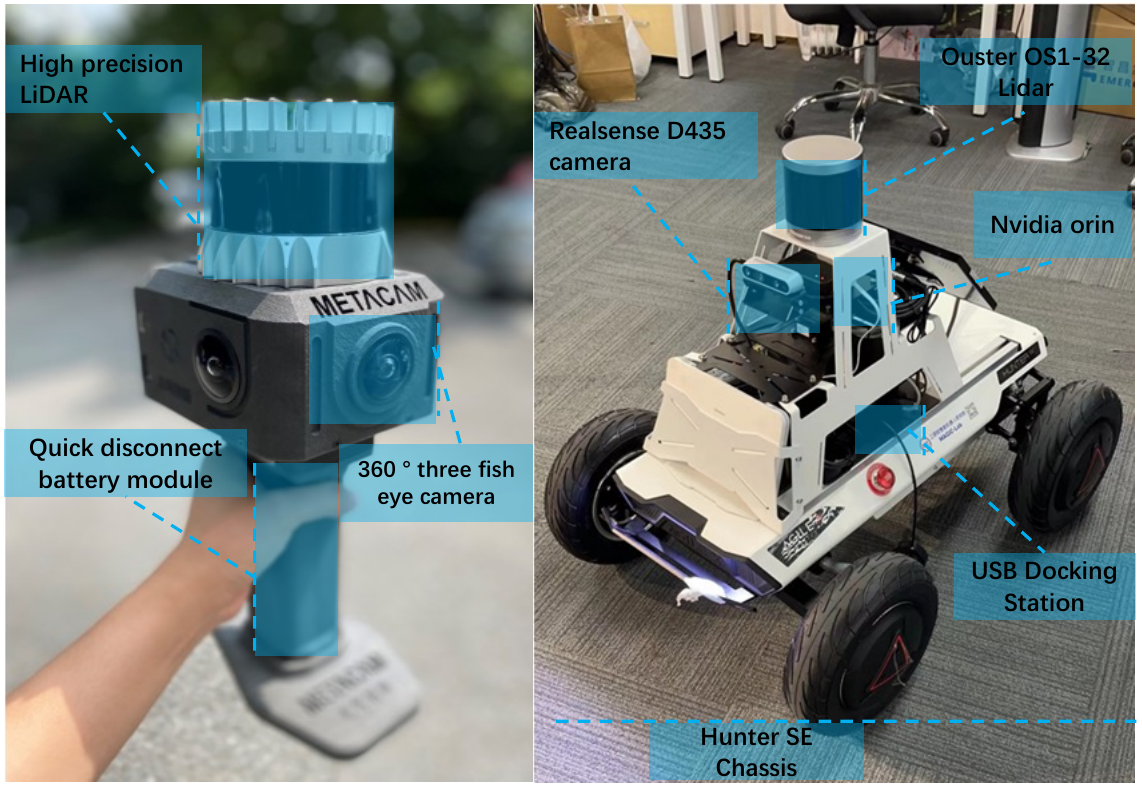}
    \caption{Two devices utilized to record in-house datasets.}
    \label{fig:devices}
\end{figure}

\section{EXPERIMENT}\label{sec:exp}

To evaluate our OpenAnnotate3D system, we conducted experiments on both public benchmark and in-house multi-modal datasets. 

\subsection{Implementation Details}

For the LLM interpreter module in our system, we utilized $\mathtt{gpt}\text{-}\mathtt{3.5}\text{-}\mathtt{turbo}$ API from OpenAI~\cite{openai} and Langchain API from Langchain~\cite{LangChain}. 
% The prompt template is displayed in Fig. \ref{fig:prompt}. 
In the promptable vision module, we integrate two off-the-shelf pre-trained models, Grounding DINO~\cite{liu2023grounding} and SAM~\cite{kirillov2023segment}, without any training or finetuning. 
Both of these models are robust foundational models that support vision-language inputs. 
Given user-provided texts and 2D images, we employ Grounding DINO to generate 2D proposal bounding boxes for images based on text feature matching. 
Subsequently, these bounding boxes are fed to the prompt encoder of SAM, serving as segmentation prompts.
SAM generates the final 2D mask based on these bounding boxes. 
It is worth noting that Grounding DINO and SAM are plug-and-play components and can be replaced with other models that have similar functionalities.
All experiments are implemented on a single machine with a GTX-4090.

\begin{table*}[!t]
    \centering
    \caption{Evaluation of 3D semantic segmentation results on selected SemanticKITTI \textit{val} subset.}
    {
        \begin{footnotesize}
        \begin{tabular}{| c || c c c c c c c |}
            \hline
            \textbf{Annotation Methods} & road & car & person & vegetation & building & pole & motorcycle \\
            \hline\hline         
            Junior User IOU(\%) & 88.6 & 63.1 & 35.4 & 67.2 & 59.1 & 18.1 & 59.8 \\
            \hline
            Senior User IOU(\%) & 91.5 & 95.3 & 67.8 & 69.9 & 88.1 & 45.3 & 84.7 \\
            \hline
            OpenAnnotate3D IOU(\%) & 94.2 & 92.3 & 75.3 & 81.4 & 85.7 & 58.2 & 93.8 \\
            \hline
            OpenAnnotate3D w.o. spatio-temporal fusion(\%) & 94.2 & 87.4 (-4.9) & 72.3 (-3.0) & 81.4 & 85.7 & 58.2 & 88.5 (-5.3) \\
            \hline                                              
        \end{tabular}
        \end{footnotesize}
    }
    \label{tab:exp_semantickitti_results}
    \vspace{-0.3cm}
\end{table*}

\begin{table*}[!t]
    \centering
    \caption{3D semantic segmentation time costs on selected SemanticKITTI \textit{val} subset (30 frames).}
    {
        \begin{footnotesize}
        \begin{tabular}{| c || c c c c c c c |}
            \hline
            \textbf{Annotation Methods} & road & car & person & vegetation & building & pole & motorcycle \\
            \hline\hline
            Junior User (Sec) & 168 & 110 & 95 & 226 & 183 & 75 & 134 \\
            \hline            
            Senior User (Sec) & 152 & 98 & 87 & 200 & 162 & 65 & 120 \\
            \hline            
            OpenAnnotate3D (Sec) & 2 & 2 & 2 & 2 & 2 & 2 & 2 \\
            \hline                       
        \end{tabular}
        \end{footnotesize}
    }
    \label{tab:exp_semantickitti_time}
    \vspace{-0.5cm}
\end{table*}

We primarily conduct experiments on two types of data:
\begin{itemize}
    \item \textbf{Public Dataset} To assess the labeling accuracy, we use our OpenAnnotate3D system to generate annotations on SemanticKITTI~\cite{behley2019semantickitti} and compare the auto-labeling results with ground truth annotations provided by SemanticKITTI.
    SemanticKITTI was collected using a 3D LiDAR and dual cameras, capturing 3D point clouds and RGB images from both left and right perspectives. 
    Since our annotation system relies on 2D images to annotate 3D data, when evaluating it on SemanticKITTI, we choose the left-view images and filter out the point clouds that were not covered by the left-view images in the 3D scene. 
    In addition, SemanticKITTI consists of 22 sequences, each containing thousands of frames. 
    For ease of comparison, we only selected the 08 \textit{val} sequence as the evaluation subset.
    \item \textbf{In-house Dataset} Public datasets only contain closed-set annotations of driving scenes, which are monotonous.
    To comprehensively evaluate open-vocabulary reasoning, we record a series of complex open-scene data using two devices. 
    The first device is a handheld 3D reality scanner called Metacam, which is equipped with a 32-line LiDAR and three fish-eye cameras.
    The LiDAR has a field of view of 45° × 360°, while the cameras output 4032$\times$3040~px full-color images.
    In addition, we also utilized an autonomous grounded vehicle with a 32-line LiDAR and a 640$\times$480 camera. The two devices are shown in Fig.~\ref{fig:devices}.
\end{itemize}

\subsection{Metrics}
To evaluate the annotation performance of OpenAnnotate3D on the 3D semantic segmentation task of the SemanticKITTI dataset, we employed IoU (intersection-over-union) for each class as the evaluation metric, i.e.,
% \begin{equation}
%     \text{mIOU} = \frac{1}{C}\sum^{C}_{c=1}\frac{TP_c}{TP_c+FP_c+FN_c}
% \end{equation}
\begin{equation}
    \text{IOU}_c = \frac{TP_c}{TP_c+FP_c+FN_c}
\end{equation}
where $TP_c$, $FP_c$, and $FN_c$ correspond to the number of true positive, false positive, and false negative point predictions for class $c$. % and $C$ is the number of classes.

Note that evaluating annotation results on closed-set datasets using IOU metrics is because closed-set annotations have ground truth. However, apart from annotating these closed-set objects, OpenAnnotate3D is capable of labeling various open-set objects, which is not reflected in the quantitative experiments due to the lack of ground truth.

\subsection{Quantitative Analysis of Accuracy and Efficiency}

Apart from labeling accuracy, we also evaluate labeling efficiency. To this end, we recruited two human annotators (one junior and one senior) who had received training in the annotating process. In the baseline comparison experiment, we manually and automatically annotated 10 objects for 30 frames from SemanticKITTI, respectively. 

OpenAnnotate3D supports a manual fine-tuning interface for users. In practice, we find that through slight manual corrections based on OpenAnnotate3D, the whole system is even more powerful thanks to the iterative process in the LLM interpreter and spatio-temporal fusion and correction. However, in this part of the experiment, we only tested the automatic annotation components of OpenAnnotate3D to represent pure auto-labeling accuracy. 

We record the precision of the annotations compared to the ground truth and the time taken by different annotators to complete the tasks. The annotation results are shown in Tab. \ref{tab:exp_semantickitti_results}, where we present the IoU for each category. Especially for objects with complex shapes like ``\textit{person}'', ``\textit{vegetation}'', or relatively small objects like ``\textit{pole}'', even senior human annotators achieve IoU of only 67.8\%, 69.9\%, and 45.3\%, respectively. In contrast, our OpenAnnotate achieves IoU of 75.3\%, 81.4\%, and 58.2\%, respectively, without any manual fine-tuning. For objects that are challenging for the human eye to precisely identify, our automatic system demonstrates a more pronounced advantage. The time costs are shown in Tab. \ref{tab:exp_semantickitti_time}.
As we can see, our OpenAnnotate3D incurs significantly lower time consumption compared to manual annotation, especially for objects with irregular shapes and large areas such as ``\textit{vegetation}'' and ``\textit{motorcycle}''. 
Furthermore, our OpenAnnotate3D, with consistent program execution speed (dependent mainly on GPU performance), can be quantified in terms of time. 
In contrast, manual annotation not only exhibits low efficiency but also varies among different users on their level of expertise.

\subsection{Qualitative Analysis of Open Vocabulary Reasoning}

In addition, as shown in Fig. \ref{fig:visualization}, we further demonstrate the annotation capabilities of our OpenAnnotate3D on real-world scene data. 
Our annotation system not only consistently and automatically annotates several common closed-set objects such as ``\textit{bicycle}'', ``\textit{person}'', ``\textit{building}'', and ``\textit{motorcycle}'', but also accurately identifies numerous open-vocabulary objects that were not previously annotated in the closed-set dataset. 
These open-vocabulary objects include ``\textit{balloon}'', ``\textit{knapsack}'', ``\textit{trunk box}, as well as long descriptions like ``\textit{the person carrying the suitcase}''. 
These examples highlight the powerful open-vocabulary annotation capabilities of our annotation system.

\subsection{Ablation Studies}
We also conduct an ablation study to evaluate the automatic correction function of our OpenAnnotate3D, namely the spatio-temporal confusion and correction module.
Using the same setup as the previous experiments, we first allowed OpenAnnotate3D to perform automatic correction. 
Then, we conducted another round of annotation with the automatic correction disabled. 
The results are shown in rows 3 and 4 of Tab. \ref{tab:exp_semantickitti_results}. 
It can be observed that after undergoing automatic correction with the spatio-temporal confusion module, the annotation system's precision is further improved, especially for moving objects like ``\textit{motorcycle}'' and ``\textit{car}''. 

\subsection{Limitations}
Our OpenAnnotate3D tool still exhibits some degree of dependency on user inputs. For prompts that are ambiguous or overly abstract, such as "other-vehicle," the annotation capabilities of the tool may be somewhat limited.
Additionally, the performance of OpenAnnotate3D is subject to hardware-specific parameters, including camera resolution, frame rate, and laser scanner resolution. In cases where camera resolution is sub-optimal, annotation results for distant or unclear objects may not meet the desired standards.

\section{CONCLUSION}

In this paper, we propose OpenAnnotate3D, an open-source open-vocabulary auto-labeling system for multi-modal 3D data, which includes an LLM-based interpreter module, a promptable vision module, and a spatial-temporal 3D auto-labeling process.
OpenAnnotate3D integrates the chain-of-thought capabilities of Large Language Models (LLMs) and the cross-modality capabilities of vision-language models. %With the LLM-based interpreter, the system can understand high-level user commands in a robust way. Moreover, through spatio-temporal fusion and correction, the annotation tool attains good auto-labeling accuracy.
To the best of our knowledge, OpenAnnotate3D is one of the pioneering works for open-vocabulary multi-modal 3D auto-labeling.

%OpenAnnotate3D leverages Large Language Models (LLMs) and promptable vision models to generate accurate 2D and 3D scene annotations.
%By following structured principles and using LLMs, we create well-structured textual descriptions for arbitrary objects in the wild to support open-vocabulary annotation. 
%Our experiments on benchmark datasets validate OpenAnnotate3D's efficiency. 
%Despite a slight drop in performance compared to manual annotation, our system significantly reduces costs and time. 
%This work showcases LLMs' potential in open-vocabulary annotation and lays the foundation for future advancements. 
%Our OpenAnnotate3D emerges as a primary solution in the evolving realm of autonomous vehicles and robotics, helping build a closed-loop data utilization for future applications.

%%%%%%%%%% APPENDIX

%%%%%%%%%% REFERENCES

\bibliographystyle{IEEEtran}
%\bibliography{root}

\end{document}